\documentclass[runningheads]{llncs}
\pdfoutput=1
\usepackage{graphicx}
\usepackage{comment}
\usepackage{amsmath,amssymb} 
\usepackage{bm}
\usepackage{color}
\usepackage{subcaption}
\DeclareMathOperator*{\argmin}{arg\,min} 

\usepackage{array}
\newcolumntype{P}[1]{>{\centering\arraybackslash}p{#1}}

\begin{document}
\pagestyle{headings}
\mainmatter
\def\ECCVSubNumber{5893}  

\title{Differentiable Programming for Hyperspectral Unmixing using a Physics-based Dispersion Model}

\titlerunning{Differentiable Programming for Hyperspectral Unmixing}
%
\author{John Janiczek\inst{1} \and
Parth Thaker\inst{1} \and
Gautam Dasarathy\inst{1}\and
Christopher Edwards\inst{2} \and Philip Christensen\inst{1} \and Suren Jayasuriya\inst{1}}
\authorrunning{J. Janiczek et al.}
%
\institute{Arizona State University\and Northern Arizona University}
\maketitle

\begin{abstract}

Hyperspectral unmixing is an important remote sensing task with applications including material identification and analysis. Characteristic spectral features make many pure materials identifiable from their visible-to-infrared spectra, but quantifying their presence within a mixture is a challenging task due to nonlinearities and factors of variation. In this paper, spectral variation is considered from a physics-based approach and incorporated into an end-to-end spectral unmixing algorithm via differentiable programming. The dispersion model is introduced to simulate realistic spectral variation, and an efficient method to fit the parameters is presented. Then, this dispersion model is utilized as a generative model within an analysis-by-synthesis spectral unmixing algorithm. Further, a technique for inverse rendering using a convolutional neural network to predict parameters of the generative model is introduced to enhance performance and speed when training data is available. Results achieve state-of-the-art on both infrared and visible-to-near-infrared (VNIR) datasets, and show promise for the synergy between physics-based models and deep learning in hyperspectral unmixing in the future.

\end{abstract}

\section{Introduction}

Hyperspectral imaging is a method of imaging where light radiance is densely sampled at multiple wavelengths. Increasing spectral resolution beyond a traditional camera's red, green, and blue spectral bands typically requires more expensive detectors, optics, and/or lowered spatial resolution. However, hyperspectral imaging has demonstrated its utility in computer vision, biomedical imaging, and remote sensing~\cite{lu2014medical,chang2003hyperspectral}. In particular, spectral information is critically important for understanding material reflectance and emission properties, important for recognizing materials. 

Spectral unmixing is a specific task within hyperspectral imaging with application to many land classification problems related to ecology, hydrology, and mineralogy~\cite{heylen2014review,keshava2002spectral}. It is particularly useful for analyzing aerial images from aircraft or spacecraft to map the abundance of materials in a region of interest. While pure materials have characteristic spectral features, mixtures require algorithms to identify and quantify material presence.


A common model for this problem is linear mixing, which assumes electromagnetic waves produced from pure materials combine linearly and are scaled by the material abundance. Mathematically this is expressed as $\mathbf{b} = \mathbf{A}\mathbf{x} + \bm{\eta}$
where $\mathbf{b}$ is the observed spectra, $\mathbf{A}$ is a matrix whose columns are the pure material spectra, $\bm{\eta}$ is the measurement noise, and $\mathbf{x}$ is the abundance of each pure material. The model assumes that the pure material spectra, referred to as endmember spectra, is known before-hand. Nonlinear effects are known to occur when photons interact with multiple materials within a scene for which we refer readers to the review by Heylen et al. for techniques to account for the non-linear mixing~\cite{heylen2014review}.

\begin{figure}
    \centering
    \includegraphics[width= \textwidth]{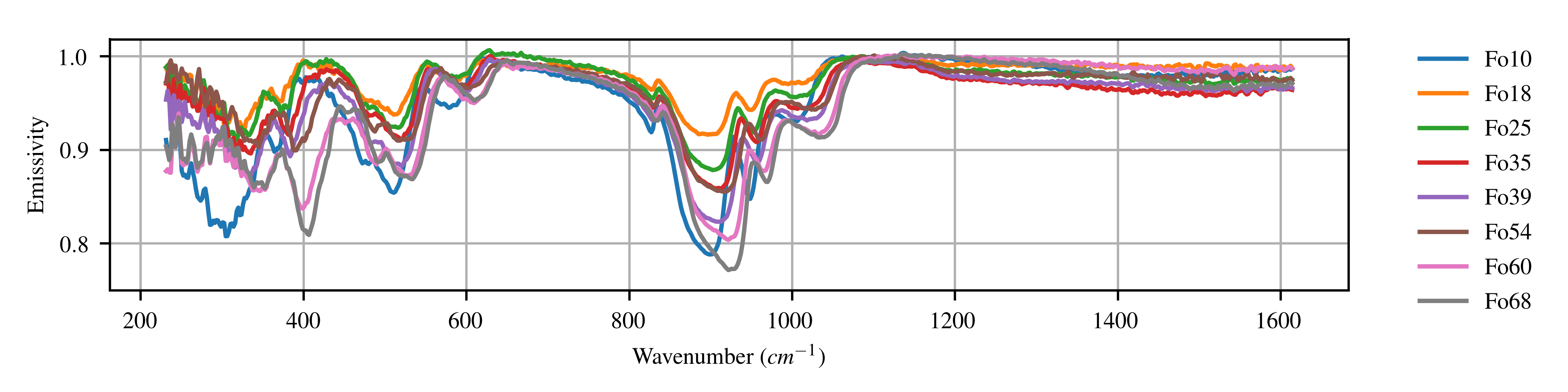}
    \caption{\textbf{Endmember Variation:} Several spectra of olivine are plotted to demonstrate it's spectral variability. The olivine mineral is a solid solution with continuous compositional variation of Iron (Fe$_2$) and Magnesium (Mg$_2$) bonds. This ratio of bonds (indexed by the Fo number), causes absorption bands to shift in frequency and strength}
    \label{fig:endmemberVariation}
\end{figure}

A key challenge that affects both linear and nonlinear mixing models is that pure materials have an inherent variability in their spectral signatures, and thus cannot be represented by a single characteristic spectrum. Spectral variability of endmembers is caused by subtle absorption band differences due to factors such as different grain sizes~\cite{moersch1995thermal,salisbury1993thermal,ramsey1998mineral,ramsey2000} or differing ratios of molecular bonds~\cite{burns1970crystal,sunshine1998determining} as shown in Figure~\ref{fig:endmemberVariation}. Since variability causes significant errors in unmixing algorithms, it is an active area of research~\cite{zare2013endmember,du2014spatial,zhou2018gaussian}. 

Recently, differentiable programming has become a popular research area due to its potential to bridge gaps between physics-based and machine learning-based techniques for computer vision and graphics~\cite{wang2018backpropagation,gkioulekas2016evaluation,azinovic2019inverse}. Our key insight is to leverage differentiable programming by modelling the variation of spectra with a physics-based dispersion model, and incorporating this differentiable model into an end-to-end spectral unmixing algorithm. Such an approach has the capacity to unmix scenes with a large amount of variability, while constraining the predictions to be physically plausible. These physically plausible variations of endmember spectra also provide additional science data as the variation of absorption bands can reveal properties about the composition and history of the material. To our knowledge, we are the first to use a generative physics model to account for spectral variability in an unmixing algorithm.
\\
\\
\textbf{Contributions:} Our specific contributions in this paper are the following:
\begin{itemize}
    \item We introduce a physics-based dispersion model (first presented in~\cite{spitzer1961infrared,larkin2017infrared,wenrich1996optical}) to generate and render spectral variation for various pure materials. We provide an efficient optimization method via gradient descent to find dispersion model parameters for this spectral variation. 
    \item We incorporate this dispersion model into an end-to-end spectral unmixing algorithm utilizing differentiable programming to perform analysis-by-synthesis optimization. Analysis-by-synthesis is solved via alternating minimization optimization and requires no training data. 
    \item We further design an inverse rendering algorithm consisting of a convolutional neural network to jointly estimate dispersion model parameters and mineral abundances for spectral unmixing. This method requires training data, but is computationally efficient at test time and outperforms analysis-by-synthesis and other state-of-the-art methods. 
\end{itemize}

We provide extensive analysis of our proposed methods with respect to noise and convergence criteria. To validate our contributions, we test on both synthetic and real datasets using hyperspectral observations in the visible and near infrared (VNIR), and mid to far infrared (IR). The datasets also span three different environments from laboratory, aircraft, and satellite based spectrometers. Our methods achieve state-of-the-art across all datasets, and we compare against several baselines from literature. Our code is openly available and accessible here: \textcolor{blue}{\url{https://github.com/johnjaniczek/InfraRender}}. We hope this work inspires more fusion between physics models and machine learning for hyperspectral imaging and computer vision more generally in the future.

\section{Related Work}
\label{sec:rw}


\textbf{Optimization-based Approaches.} Standard optimization techniques for linear unmixing include projection, non-negative least squares, weighted least squares, and interior point methods~\cite{ramsey1998mineral,heinz2001fully,rogers2008mineralogical,chouzenoux2014fast,goudge2015integrating}. Further, sparsity-based optimization can improve abundance prediction~\cite{zhang2018spectral,chen2013sparse}. However, most optimization have not leveraged physics priors as we do in our model. 

\textbf{Spectral Variability.} Spectral variability has been a topic of recent interest~\cite{zare2013endmember,borsoi2020spectral}. One approach is to augment $\mathbf{A}$ with multiple variations or spectra for each endmember. To do this, multiple endmember spectral mixture analysis (MESMA)~\cite{roberts1998mapping} and multiple-endmember linear spectral unmixing (MELSUM)~\cite{combe2008analysis} both require labeled data of the spectral variation for each endmember. In contrast, unsupervised techniques learn endmember sets from unlabelled hyperspectral images, including semi-automated techniques~\cite{bateson2000endmember}, k-means clustering~\cite{bateson2000endmember}, and the sparsity promoting iterated constrained endmember algorithm (SPICE)~\cite{zare2007sparsity,zare2008hyperspectral} which simultaneously finds endmember sets while unmixing for material abundances. These techniques are limited by the amount of sets in the endmember library, and computational complexity increases with more additions. Our method by contrast finds an efficient parameter set to physically model the spectral variation. 

Another category of endmember variability techniques models the endmember spectral variation as samples from a multivariate distribution $\mathbf{P}(\mathbf{e}|\theta)$ where $\mathbf{e}$ is the endmember spectra, and $\theta$ are the distribution parameters. Common statistical distributions proposed include the normal compositional model~\cite{stein2003application}, Gaussian mixture models~\cite{zhou2018gaussian}, and the beta compositional model~\cite{du2014spatial}. These distribution models have large capacity to model spectral variations, however sometimes they can render endmember spectra that are not physically realistic.


\textbf{Deep Learning for Hyperspectral Classification and Unmixing.} Deep learning has recently improved many hyperspectral imaging tasks~\cite{zhang2016deep,li2019deep}. In particular, networks process hyperspectral pixel vectors using both deep belief networks~\cite{liu2016active} and CNNs~\cite{hu2015deep,chen2016deep,yang2018hyperspectral}. For spatial hyperspectral data, CNNs~\cite{cheng2018exploring}, joint spectral-spatial feature extraction~\cite{zhao2016spectral}, and 3D CNNs~\cite{li2017spectral} are used. All these methods require large hyperspectral datasets that are annotated correctly. One of our methods uses analysis-by-synthesis and differentiable programming to avoid low training data issues, but our technique can be made complementary to deep learning architectures as we show in our inverse rendering CNN.


\textbf{Differentiable Programming and Rendering.} Differentiable programming refers to the paradigm of writing algorithms which can be fully differentiated end-to-end using automatic differentiation for any parameter~\cite{wang2018backpropagation,wang2019demystifying,baydin2017automatic}. This has been applied for audio~\cite{Engel2020DDSP} and 3D geometry processing~\cite{ravi2020pytorch3d}. In graphics, differentiable rendering has improved ray tracing~\cite{Li:2018:DMC,NimierDavidVicini2019Mitsuba2,Loubet2019Reparameterizing,zhang2019differential}, solved analysis-by-synthesis problems in volumetric scattering~\cite{gkioulekas2013inverse,gkioulekas2016evaluation}, estimated reflectance and lighting~\cite{azinovic2019inverse}, and performed 3D reconstruction~\cite{tsai2019beyond}. In our paper, we write a forward imaging model utilizing the physics of dispersion in spectral variation to allow our pipeline to be differentiable end-to-end.

\section{Method} 

Our approach to hyperspectral unmixing features two main components: (1) use of a physically-accurate dispersion model for pure endmember spectra, and (2) a differentiable programming pipeline to perform spectral unmixing. This approach has synergistic benefits of leveraging prior domain knowledge while learning from data. Our first algorithm solves spectral unmixing in a self-supervised fashion using analysis-by-synthesis optimization with the dispersion model as the synthesis step. Further, we show how inverse rendering via a convolutional neural network (CNN) can learn parameters of this model to help speed up our end-to-end pipeline and improves performance when training data is available. 


\subsection{Dispersion Model}
\label{sec:dispersionModel}
We first describe the dispersion model for generating endmember spectra. \textbf{Endmember and/or endmember spectra} is what we call the spectral curve for emissivity $\epsilon$ as a function of wavenumber $\omega$. Each pure material has a characteristic endmember spectrum, although spectra can vary, which is the problem we are trying to solve/disambiguate. 
\newcommand{\epsmeasured}{\varepsilon^{\rm measured}}
\newcommand{\epsmodel}{\varepsilon^{\rm model}}
Let $\epsmeasured(\omega)$ be endmember spectra we have measured, typically in a lab or in the field, whose emissivity is sampled at different wavenumbers:  
$\left[ \epsmeasured(\omega_1), \cdots \epsmeasured(\omega_N)\right]^T$. Our goal is to propose a model $\epsmodel(\bm{\Lambda};\omega)$ with parameters $\bm{\Lambda}$ such that the following loss is minimized:
$ L(\bm{\Lambda}) = \sum_{i=1}^{N} \left(\epsmeasured(\omega_i) - \epsmodel(\bm{\Lambda}; \omega_i)\right)^2. $
That is, we fit the model emissivity of an endmember spectrum to the measured spectrum. In practice, we need to add regularization and constraints to this endmember loss for better fitting which we describe after the derivation of the dispersion model. 

\begin{figure}
    \centering
    \includegraphics[width= 0.85\textwidth]{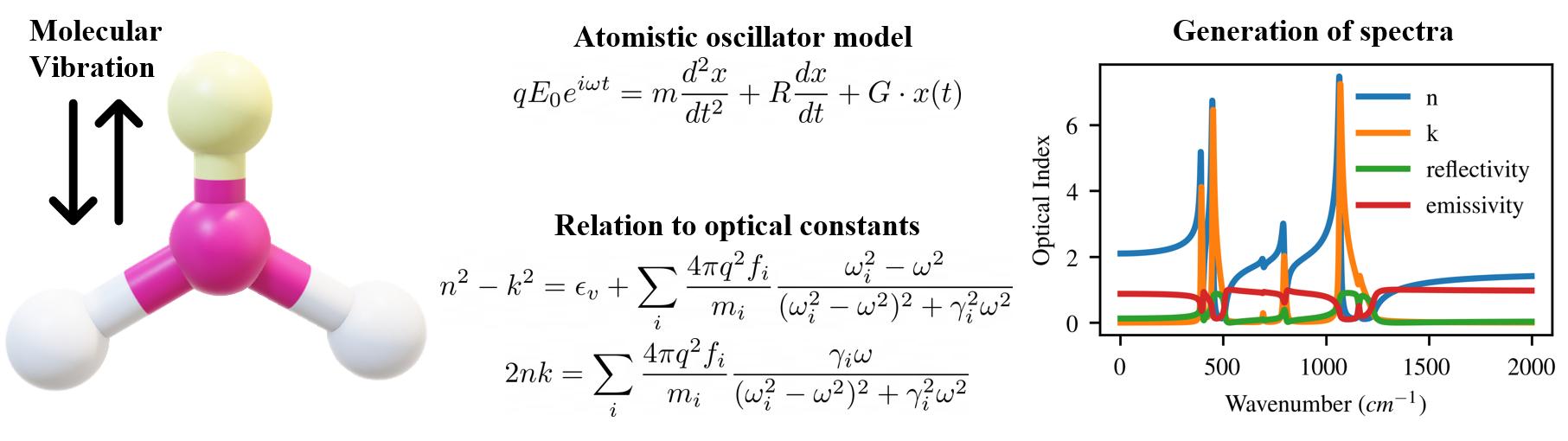}
    \caption{\textbf{Dispersion Model Concept Figure:} The insight of the dispersion model is that optical properties can be related to molecular structure through first principles via an atomistic oscillator model. We use this generative model for the formation of spectral variation in our spectral unmixing algorithm}
    \label{fig:dispersionmodel}
\end{figure}

\indent \textbf{Derivation of the Dispersion model:} Our model of endmember spectra is derived from an atomistic oscillator driven by electromagnetic waves impinging on the molecular structure of the pure material~\cite{spitzer1961infrared,larkin2017infrared}. In Figure~\ref{fig:dispersionmodel}, we show a conceptual diagram of this model, and how it generates emissivity curves as a function of wavelength. For the full derivation of the model from first principles, we refer the reader to Appendix A in the supplementary material. Instead, we outline the model below based on the equations derived from that analysis.

Let $\bm{\Lambda} = \left[ \bm{\rho}, \bm{\omega_o}, \bm{\gamma},\bm{\epsilon_r} \right]$ be a matrix of parameters, where $\bm{\rho},\bm{\omega_0}, \bm{\gamma}, \bm{\epsilon_r}  \in \mathbb{R}^K$ and $K$ is a model hyperparameter corresponding to the number of distinct mass-spring equations used to model the emissivity. $\bm{\rho}$ is the band strength, $\bm{\omega_o}$ is the resonant frequency, $\bm{\gamma}$ is the frictional force (dampening coefficient), and $\bm{\epsilon_r}$ is relative dielectric permeability. Please see the supplemental material for the physical significance of these parameters to the atomistic oscillator model, and their control over the shape of spectral absorption bands. Note: usually $\bm{\epsilon_r}$ is a constant vector which does not vary with $K$. Thus $\bm{\Lambda} \in \mathbb{R}^{K\times 4}.$ The refractive index terms $n,k$ are given as follows~\cite{spitzer1961infrared,larkin2017infrared}:
\begin{equation}
    n(\bm{\Lambda};\omega) = \sqrt{\frac{\theta + b}{2}}, \quad k(\bm{\Lambda};\omega) = \frac{\phi}{n(\bm{\Lambda};\omega)},
\end{equation}
where the expressions for $\theta, b, \phi$ are given as follows:
\begin{equation}
    \theta = \epsilon_r + \sum_{k=1}^{K} 4\pi \rho_k \omega_{0_k}^2 \frac{(\omega_{0_k}^2 - \omega^2)}{(\omega_{0_k}^2 - \omega^2)^2 + \gamma_k^2 \omega_{0_k}^2 \omega^2},
\end{equation}
\begin{equation}
    b = \sqrt{\theta^2 + 4\phi^2},\quad \phi = \sum_{k=1}^{K} 2\pi \rho_k \omega_{0_k}^2  \frac{\gamma_k \omega_{0_k} \omega}{(\omega_{0_k}^2 - \omega^2)^2 + \gamma_k^2 \omega_{0_k}^2 \omega^2}.
\end{equation}
We note that subscript $k$ denotes the k-th coordinate of the corresponding vector. Also there is another useful relation (derived in Appendix A) that $n^2 - k^2 = \theta, nk = \phi.$ We then define the complex refractive index as $
\hat{n}(\bm{\Lambda};\omega) = n(\bm{\Lambda};\omega) - i\cdot k(\bm{\Lambda};\omega),
$
where $i = \sqrt{-1}$ is the imaginary number. Hence, we can calculate the emissivity as follows:
\begin{equation}
\label{eq:kirchoff}
\epsilon(\bm{\Lambda};\omega) = 1 - R(\bm{\Lambda};\omega), \qquad\mbox{ where }
R(\bm{\Lambda};\omega) = \left|\frac{\hat{n}(\bm{\Lambda};\omega) - 1}{\hat{n}(\bm{\Lambda};\omega)+1} \right| ^2.
\end{equation}

When considering minerals, we introduce $M \in \mathbb{N}$, the number of optical axes of symmetry in crystal structures, (eg. 2 axes of symmetry in quartz~\cite{spitzer1961infrared,wenrich1996optical}), to define the full model:
\begin{equation}
    \epsmodel(\bm{\Lambda};\omega) = \sum_{m=1}^{M} \alpha_{m}\cdot \epsilon(\bm{\Lambda_m};\omega)\quad \text{such that} \sum_{m=1}^{M}\alpha_m = 1, \alpha_m \geq 0,
\end{equation}

\noindent where we use a different parameter matrix $\bm{\Lambda_m}$ and weight $\alpha_m$ for each optical axis of symmetry. 

The dispersion model has been primarily used to analyze optical properties of materials to determine $n$ and $k$, which then can be subsequently applied to optical models like radiative transfer~\cite{spitzer1961infrared,wenrich1996optical}. After $n$ and $k$ are found, spectra such as reflectance, emissivity, and transmissivity can be generated. In particular, we notice that fine-grained control of the dispersion model parameters can realistically render spectral variation that occurs in hyperspectral data. Our contribution is to leverage these properties in a differentiable programming pipeline for spectral unmixing. 

\textbf{Endmember Fitting.} Using the dispersion model presented above, we want to robustly estimate the model parameters to fit the spectra $\epsmeasured{}$ captured in a lab or in the field. To fit the model, we wish to perform gradient descent to efficiently find these parameters. Using chain rule on the loss function, we see that $\frac{\partial L}{\partial \bm{\Lambda}_{ij}} = \frac{\partial L}{\partial \epsmodel} \frac{\partial \epsmodel}{\partial \bm{\Lambda}_{ij} },
$
where $(ij)$ corresponds to that element of the parameter matrix, and all expressions are scalars once the coordinate is specified. While the partial derivatives can be calculated explicitly via symbolic toolboxes, the resulting expressions are too long to be presented here. For simplicity and ease of use, we use the autograd function~\cite{baydin2017automatic,paszke2017automatic} in PyTorch~\cite{paszke2019pytorch} to automatically compute derivatives for our model as we are performing backpropagation. 

One main challenge in performing endmember fitting is that the dispersion model is not an injective function, and hence is typically not identifiable, that is more than one $\bm{\Lambda}$ can result in the same fit.  This can be solved, in-part, through regularization to enforce sparsity, especially since a preference for fewer dispersion parameters has been suggested in the literature~\cite{spitzer1961infrared,wenrich1996optical}. In our implementation, we initialize our model with $K = 50$ rows of the parameter matrix. Since the parameter $\rho$ controls the strength of the absorption band, small values of $\rho$ do not contribute much energy to the spectra (unnecessary absorption bands), and can be pruned. After performing sparse regression by penalizing the $L_1$ norm of $\rho$, $K$ is typically around 10-15 in our experiments.

Thus, our modified sparse regression problem may be written as 
\begin{equation}
\label{eq:endmemberloss}
\argmin_{ \bm{\Lambda}_{\rm min} \leq\bm{\Lambda} \leq \bm{\Lambda}_{\rm max}}  \sum_{i=1}^{N} \left(\epsilon^{real}(\omega_i) - \epsilon^{model}(\bm{\Lambda}; \omega_i)\right)^2 + \lambda_\rho ||\rho||_1,
\end{equation}
where $\bm{\Lambda}_{\rm min}$ and $\bm{\Lambda}_{\rm max}$ restrict the variation of the dispersion parameters to a plausible range. In addition, endmembers (particularly minerals) can have multiple optical axes of symmetry described by separate spectra, which has been noted in the literature~\cite{spitzer1961infrared,wenrich1996optical}. Without prior knowledge of the number of axes for every material we encounter, we run this optimization for a single and double axes, and pick the one with the lowest error. See Section ~\ref{sec:results} for results on endmember fitting and Figure~\ref{fig:endmemberfitting} for examples of modelled vs. measured spectra.

Despite the fact that this regression problem is non-convex, we solve it using gradient descent with a random initialization; this is known to converge to a local minimum with probability 1~\cite{lee2016gradient}. A global minimum is not necessary at this stage, since we use endmember fitting to provide a good initialization point for the subsequent alternating minimization procedure introduced in the next subsection. 

\subsection{Differentiable Programming for End-to-End Spectral Unmixing}
\label{sec:diffprog}

\begin{figure}
    \centering
    \includegraphics[width=0.85\textwidth]{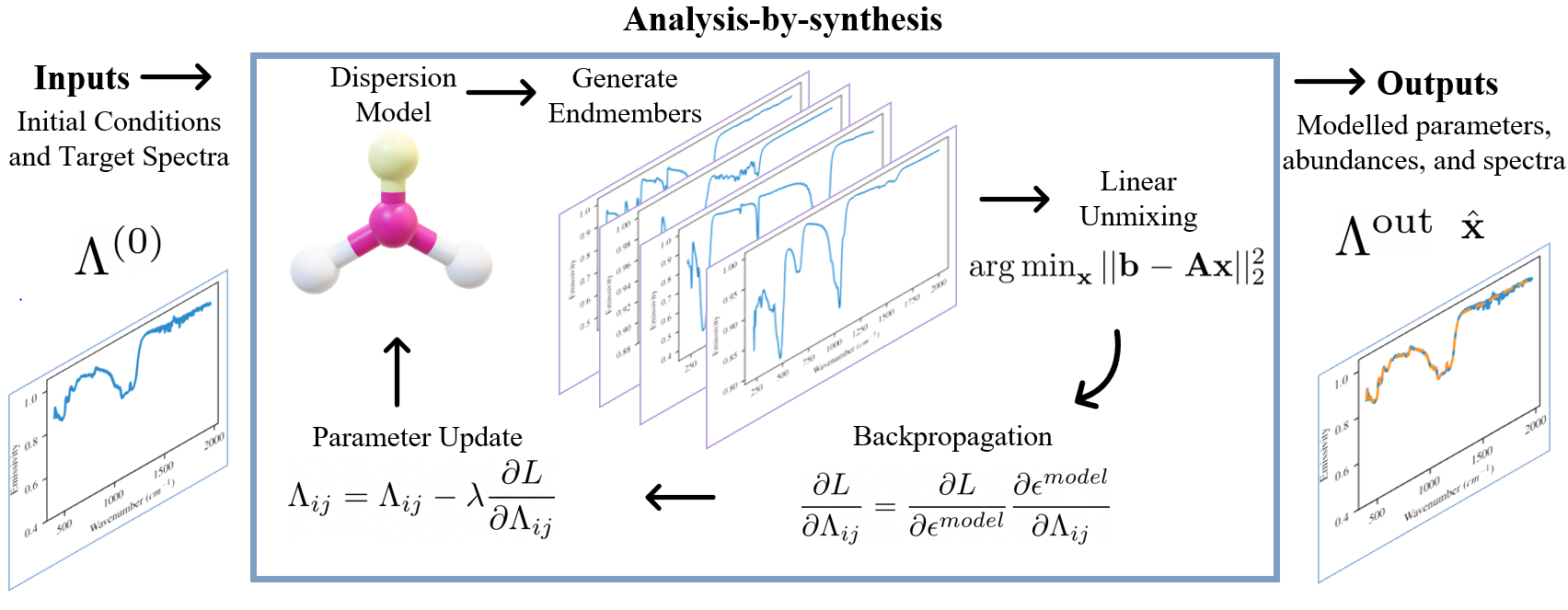}
    \caption{\textbf{Analysis-by-Synthesis:} The analysis-by-synthesis algorithm uses differentiable programming to find optimal dispersion parameters and abundances. The initial dispersion parameters and the target spectra are fed as inputs, and the algorithm alternates between optimizing the abundances in the least squares sense and updating the dispersion parameters with respect to the gradient}
    \label{fig:DiffProgramming}
\end{figure}

\textbf{Analysis-by-Synthesis Optimization.} In Figure~\ref{fig:DiffProgramming}, we show our full end-to-end spectral unmixing pipeline. Here, $\epsmodel{}(\Lambda;\omega)$, which is initially fit to $\epsmeasured{}$, is then aggregated into the columns of $\mathbf{A}$. Then, the observed spectra $\mathbf{b}$ is linearly unmixed by solving a regularized least-squares optimization:
$
\text{argmin}_{\mathbf{x}} \|\mathbf{b} - \mathbf{A}\mathbf{x}\|_2^2 + \lambda \|\mathbf{x}\|_p
$
 subject to sum-to-one and non-negativity constraints $\|\mathbf{x}\|_1 = 1, \mathbf{x} \geq 0$. Given these constraints, one cannot impose sparsity with the usual $L_1$ norm. Instead, we use the $L_p$ norm to induce sparsity for the predicted abundances; this has been proposed before for spectral unmixing~\cite{chen2013sparse}. 

The key to our pipeline is that everything is fully differentiable, and thus we can actually minimize the following equation: 
\begin{equation}
\label{eq:diffprogloss}
    \argmin_{\substack{\mathbf{x},  \bm{\Lambda} \in [\bm{\Lambda}_{min}, \bm{\Lambda}_{max}]}} \|\mathbf{b} - \mathbf{A}(\bm{\Lambda})\mathbf{x}\|_2^2 + \lambda \|\mathbf{x}\|_p \text{ such that } \|\mathbf{x}\|_1 = 1, \mathbf{x}\geq 0.
\end{equation} 
with respect to both the parameters of the dispersion model $\bm{\Lambda}$ and the abundances $\mathbf{x}$. This gives us our recipe for hyperspectral unmixing: first, perform endmember fitting to initialize $\mathbf{A}(\bm{\Lambda})$, then, solve Equation~\ref{eq:diffprogloss} in an alternating fashion for $\mathbf{x}$ and $\bm{\Lambda}$. One could also solve this equation jointly for both unknowns, however, we found that the alternating optimization was faster and converged to better results. 
\color{black}

The optimization problem established in equation~\eqref{eq:diffprogloss} is an alternating minimization problem and is unfortunately not convex~\cite{jain2017non}. One popular approach to tackle nonconvex problems is to find a good initialization point~\cite{duchi2019solving,bhojanapalli2016dropping}, and then execute a form of gradient descent. Inspired by this, we first initialize $\mathbf{A}(\bm{\Lambda})$ by performing endmember fitting using Equation~\ref{eq:endmemberloss} as described in the previous subsection. Our experiments indicate that this provides a useful initialization for our subsequent step. We then perform alternating minimization on Equation~\ref{eq:diffprogloss} for $\mathbf{x}$ and $\bm{\Lambda}$. Note that each iterate of the resulting alternating minimization involves the solution of a subproblem which has a convergence rate which depends on the condition number of the matrix $\mathbf{A}(\bm{\Lambda})$. For more details on this, we refer the reader to Appendix~C where we discuss on the properties of $\mathbf{A}(\bm{\Lambda})$ across multiple runs. 

In the ideal scenario, this initial matrix $\mathbf{A}(\bm{\Lambda})$ would consist of the endmember spectra that fully characterizes the mixed spectra $\mathbf{b}$. However, since spectra for the same material can significantly vary~\cite{moersch1995thermal,salisbury1993thermal,ramsey1998mineral,ramsey2000,burns1970crystal,sunshine1998determining} (see Figure \ref{fig:endmemberVariation}), the initialization can be slightly off and we follow up with~\eqref{eq:diffprogloss} to obtain a better fit.
Note that this optimization problem is solving for the maximum likelihood estimator under a Gaussian noise model. Our optimization technique is performing \textbf{analysis-by-synthesis}, as given a single observation $\mathbf{b}$, the dispersion model synthesizes endmember variation until a good fit is achieved.

\begin{figure}
    \centering
    \includegraphics[width=0.9\textwidth]{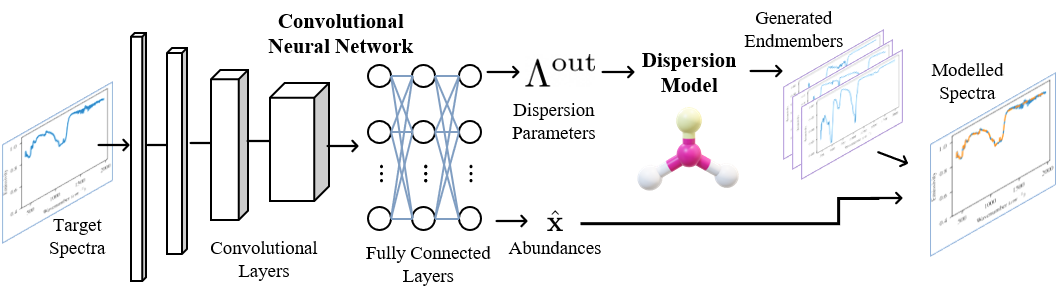}
    \caption{Inverse Rendering: A CNN is trained to ``inversely render" pixels of the hyperspectral image, by predicting both the dispersion parameters that control the spectral variability, and the abundances that control the mixing model. During training, the reconstruction error is back-propagated through the differentiable dispersion model to boost the performance of the network at making physically realistic predictions}
    \label{fig:InverseRendering}
\end{figure}

\textbf{Inverse Rendering of Dispersion Model Parameters.} The previous analysis-by-synthesis optimization does not require training data (labeled abundances in spectral mixtures) in order to perform spectral unmixing. However, there is room for even more improvement by using labeled data to help improve the parameter fitting of the model in the synthesis step. We train a CNN to predict the parameters for a generative model, known as inverse rendering in other domains~\cite{yu2019inverserendernet}. In Figure~\ref{fig:InverseRendering}, we show this inverse rendering conceptually, and how it can be fed into our differentiable programming pipeline for end-to-end spectral unmixing.

Our CNN architecture consists of convolutional layers followed by a series of fully-connected layers. We refer the reader to the supplemental material for the exact network structure and implementation details. Using a CNN for inverse rendering is significantly faster at test time as compared to the analysis-by-synthesis optimization. However, it does have a drawback of requiring training data which is unavailable for certain tasks/datasets. 

\section{Experimental Results}
\label{sec:results}

\begin{figure}
    \centering
    \includegraphics[width=0.7\textwidth]{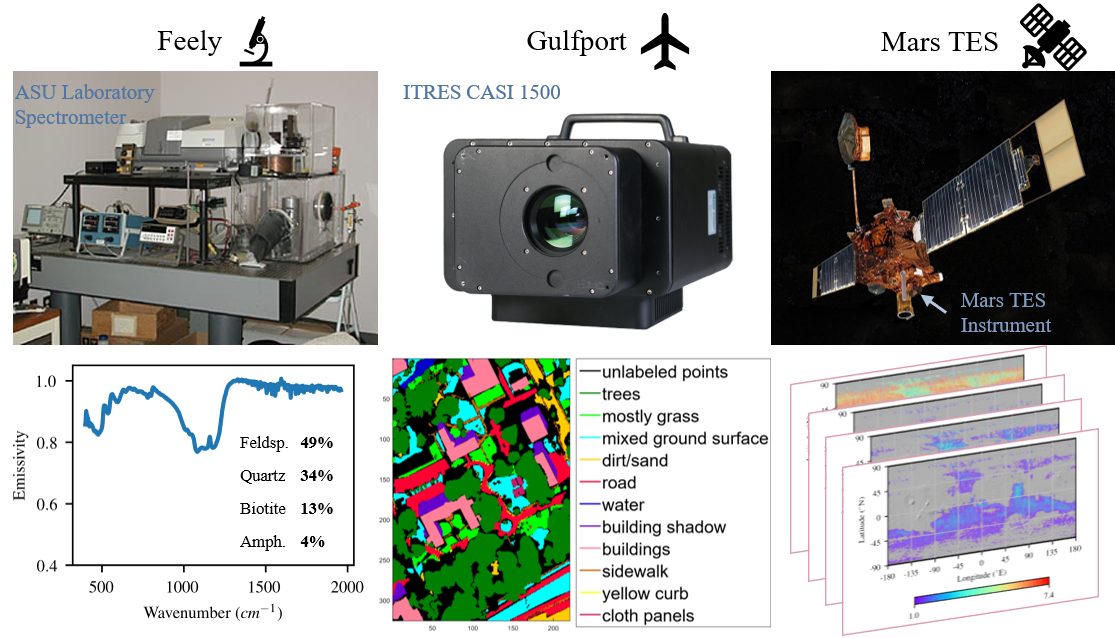}
    \caption{\textbf{Datasets:} This figure shows representative data and instrumentation for the three datasets considered in this paper. Data includes laboratory, aircraft, and satellite measurements, and ground truth ranges from detailed abundance analysis under a microscope (Feely~\cite{feely1999quantitative}) to pure pixel labels of land type for spectral classification (Gulfport~\cite{gader2013muufl}) to no ground truth for the Martian data (TES~\cite{christensen2001mars}) }
    \label{fig:datasets}
\end{figure}

\textbf{Datasets.} We utilize three separate datasets to validate our spectral unmixing algorithms. In Figure~\ref{fig:datasets}, we visually represent these datasets and their exemplar data. For specific implementation details and dataset pre-processing, please see Appendix D in the supplemental material.

\textit{Feely et al. Dataset.} We utilize 90 samples from the Feely et al. dataset~\cite{feely1999quantitative} of thermal emission spectra in the infrared for various minerals measured in the lab. Ground truth was determined via optical petrography~\cite{feely1999quantitative}, and a labeled endmember library is provided. The limited amount of data is challenging for machine learning methods, so we utilize the dispersion model to generate 50,000 additional synthetic spectra for dataset augmentation. 

\textit{Gulfport dataset.} The Gulfport dataset from Gader et al.~\cite{gader2013muufl} contains hyperspectral aerial images in the VNIR along with ground truth classification labels segmenting pixels into land types (e.g. grass, road, building). Although the dataset is for spectral classification, it can also be used to benchmark unmixing algorithms by creating synthetic mixtures of pure pixels from the Gulfport dataset with random abundances as done by~\cite{du2014spatial,zhou2018gaussian}. We perform both spectral classification (Gulfport) and unmixing (Gulfport synthetic) tasks in our testing. Both datasets are split into a train and test set (although some methods do not require training data), and the training data is augmented with 50,000 synthetically generated mixtures from the dispersion model.

One main difficulty of this dataset is the endmembers identified correspond to coarse materials such as grass and road as opposed to pure materials. Such endmembers can significantly vary across multiple pixels, but this spectral variation is not physically described by the dispersion model. To solve this problem, we utilize K-means clustering to learn examplar endmembers for each category (e.g. grass, road, etc). Then the resulting centroid endmember can be fit to the dispersion model to allow further variation such as absorption band shifts in the spectra. We found that $K = 5$ worked the best for the Gulfport dataset.


\textit{TES Martian Dataset.} The Thermal Emission Spectrometer (TES)~\cite{christensen2001mars} uses Fourier Transform Infrared Spectroscopy to measure the Martian surface. We utilize pre-processing from Bandfield et al.~\cite{bandfield2002global}, and the endmember library used by Rogers et al. to analyze Mars TES data~\cite{rogers2008mineralogical}. There is no ground truth for this dataset, as the true abundance of minerals on the Martian surface is unknown, so other metrics such as reconstruction error of the spectra are considered.

\textbf{Baselines.} We compare against several state-of-the-art baselines in the literature. The basic linear unmixing algorithm is Fully Constrained Least Squares (FCLS)~\cite{heinz2001fully} which solves least squares with sum-to-one and non-negativity constraints on the abundances. We also implement two state-of-the-art statistical methods for modelling endmember variability as distributions: the Normal Compositional Model (NCM)~\cite{stein2003application} and the Beta Compositional Model (BCM)~\cite{du2014spatial}. NCM and BCM use a Gaussian and Beta distribution respectively, perform expectation-maximization for unmixing, and require a small amount of training data to determine model parameters. 

We also compare against two state-of-the-art deep learning networks by Zhang et al.~\cite{zhang2018spectral}. The first network utilizes a 1D CNN (CNN-1D) architecture, while the second network utilizes a 3D CNN (but with 1D convolutional kernels) (CNN-3D). CNN-3D is only applicable to datasets with spatial information, and not testable on the Feely and Gulfport synthetic data. We further created a modified CNN architecture (CNN-1D Modified) to maximize the performance on our datasets by changing the loss function to MSE, removing max-pooling layers, and adding an additional fully connected layer before the output. In the supplemental material, we provide information about the parameters, network architectures, and training procedures we used for these baselines as well as details for our own methods. We also have all of our code available here: \textcolor{blue}{\url{https://github.com/johnjaniczek/InfraRender}}.

\begin{figure}
    \centering
    \includegraphics[width=0.9\textwidth]{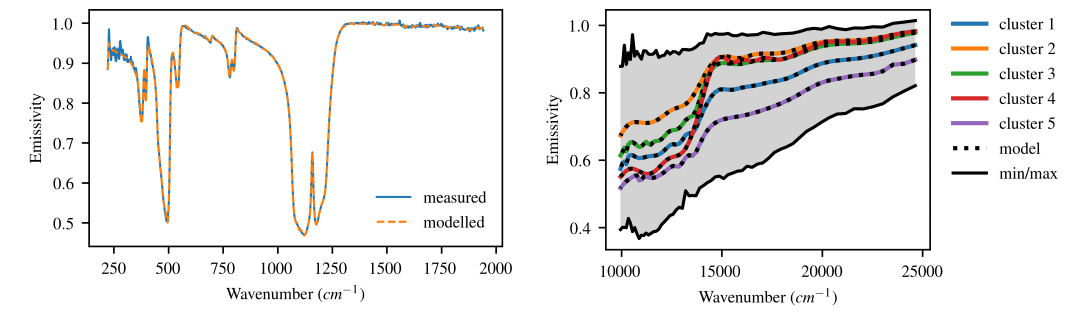}
    \caption{\textbf{Endmember Fitting:} (Left) Measured and modelled spectra for a quartz sample in the IR. (Right) Cluster centroids found for pixels labelled as grass in the Gulfport dataset, and the model fit to these centroids. Note the high fidelity of fit via the dispersion model for both these cases}
    \label{fig:endmemberfitting}
\end{figure}

\textbf{Endmember Fitting Results.} To bootstrap both the analysis-by-synthesis and inverse rendering algorithms, good initial conditions for the dispersion parameters need to be input to the model. Determining dispersion parameters typically required detailed molecular structure analysis or exhaustive parameter searching methods~\cite{spitzer1961infrared,wenrich1996optical,lee1981optical}. One main advantage of our method is that we utilize gradient descent to efficiently find parameter sets for different materials. In Appendix B of the supplemental material, we share some of these parameter sets and our insights using the dispersion model for the scientific community.

In Figure~\ref{fig:endmemberfitting}, we show qualitative results of our endmember fitting by minimizing the loss in Equation~\ref{eq:endmemberloss} using gradient descent. The reconstructed spectra achieves a low MSE with the measured spectra with an average MSE of 0.016 for the TES library, 0.0019 for the Feely library, and an MSE of 2.6e-5 on the Gulfport cluster centroids. Note that there is noise in the measurements, and so MSE is not an absolute metric of the fit to the true unknown spectra.

\textbf{Spectral Unmixing Results.} In Table~\ref{tab:mainresults}, we show results on the Feely, Gulfport, and the Gulfport synthetic mixture datasets. For Feely, the analysis-by-synthesis method achieved a MSE of 0.052, with the next closest method (NCM) achieving 0.119. Due to the Feely dataset only containing 90 test samples, the machine learning methods were trained on synthetic data which explains their lower performance as data mismatch. Thus, the low error of analysis-by-synthesis shows the utility of the dispersion model for modelling endmember variability, particularly in cases with low training data.

For the Gulfport dataset, the task was to predict the material present since the labeled data is for single coarse materials (e.g. road, grass, etc) at 100\% abundance per pixel. Here, the deep learning methods of CNNs and Inverse Rendering have the highest performance. This is expected as there exists a large amount of training data to learn from. Note that Inverse Rendering performs the best at 0.272 MSE, demonstrating that the addition of a generative dispersion model to the output of the CNN improves performance over purely learned approaches. Also note that our analysis-by-synthesis method still has relatively high performance (0.45 MSE) without using any training data at all. 

For the Gulfport synthetic mixture dataset, Inverse Rendering achieves the lowest MSE of 0.059, leveraging both physics-based modeling for spectral mixing as well as learns from available training data. The BCM and the analysis-by-synthesis methods both outperform the CNN methods, even though they do not have access to the training data. In fact, BCM even slightly outperforms the analysis-by-synthesis method, which could be because the sources of variation in this data are well-described by statistical distributions.

\begin{table}
\centering
\caption{\textbf{Results:} Table - Mean squared error of the abundance predictions vs. ground truth for Feely, Gulfport, and Gulfport synthetic datasets. The bold entries indicate top performance}
\small
\begin{tabular}{|P{1.4cm}|P{0.9cm}|P{0.9cm}|P{0.9cm}|P{1.3cm}|P{1.3cm}|P{1.3cm}|P{1.5cm}|P{1.5cm}|}
\hline

 Dataset & FCLS ~\cite{heinz2001fully}& NCM ~\cite{stein2003application} & BCM ~\cite{du2014spatial}& CNN-1D \cite{zhang2018spectral}& CNN-3D \cite{zhang2018spectral} & CNN-1D Modified & Analysis-by-synthesis & Inverse Rendering \\ \hline
Feely ~\cite{feely1999quantitative} & 0.121 & 0.119 & 0.131 & 0.469 & N/A & 0.205 & \textbf{0.052} & 0.188 \\ \hline
Gulfport ~\cite{gader2013muufl} & 0.75 & 0.799 & 0.800 & 1.000 & 0.497 & 0.297 & 0.45 & \textbf{0.272} \\ \hline
Gulfport Synthetic & 0.911 & 0.471 & 0.136 & 0.824 & N/A & 0.148 & 0.147 & \textbf{0.059} \\ \hline
\end{tabular}
\label{tab:mainresults}
\end{table}


\textit{Speed of Methods.} The additional capacity of adding statistical and physical models usually has a cost of speed in implementation. Averaged over 90 mixtures, the convergence for a single operation was FCLS - 10ms, BCM - 1.23s, NCM - 18ms, CNN - 33ms, Inverse Rendering - 39ms, and analysis-by-synthesis - 10.2s. Future work could potentially increase the speed of analysis-by-synthesis with parallel processing.

\begin{figure}
    \centering
    \includegraphics[width=\textwidth]{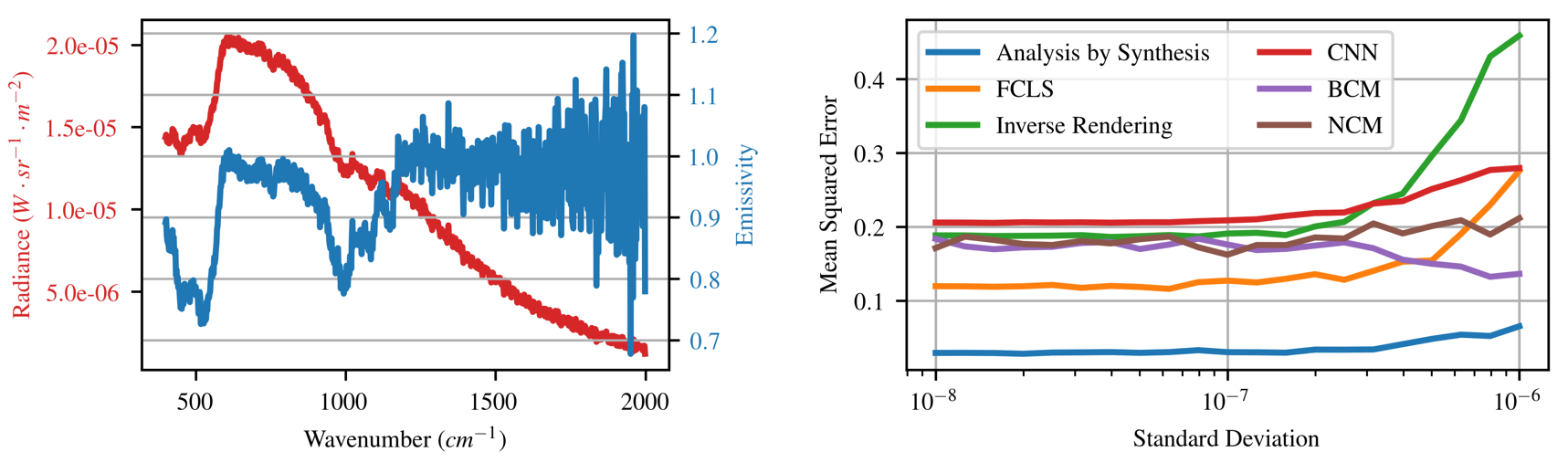}
    \caption{The left plot shows the radiance profile of a spectra perturbed by Gaussian noise and the resulting emissivity profile after separating out the blackbody radiance. The right figure shows the robustness of the algorithms to increasing amounts of noise}
    \label{fig:noisefigure}
\end{figure}

\textbf{Noise analysis.} Prior to spectral unmixing, emissivity is separated from radiance by dividing out the black-body radiation curve at the estimated temperature~\cite{ramsey1998mineral,christensen2018osiris}. In general, a Gaussian noise profile in the radiance space with variance $\sigma^2_{\text{radiance}}$ results in wavenumber dependent noise source in the emissivity space with the profile $\sigma^2(\omega) = \sigma^2_{\text{radiance}}\cdot 1/B(\omega,T)$ where $B$ is the black-body function given by Planck's law. In our noise experiments we use a black-body radiation curve for a 330K target, which is the approximate temperature the Feely dataset samples were held at. In Figure~\ref{fig:noisefigure} left, we see that the emissivity noise is higher where the radiance signal is lower.

We simulated varying the noise power to determine the methods' robustness tested on 30 samples from the Feely dataset. In Figure~\ref{fig:noisefigure}, you can see that analysis-by-synthesis still has the best performance in the presence of noise, and is relatively flat as noise increases compared to other methods. We note that statistical methods, while having higher average error, seem to be robust to increased noise as they can handle random perturbations of each spectral band statistically. CNN and Inverse Rendering methods perform the worst for high noise, as these methods were trained on data without noise. 

\begin{figure*}
\centering
\includegraphics[width =\textwidth]{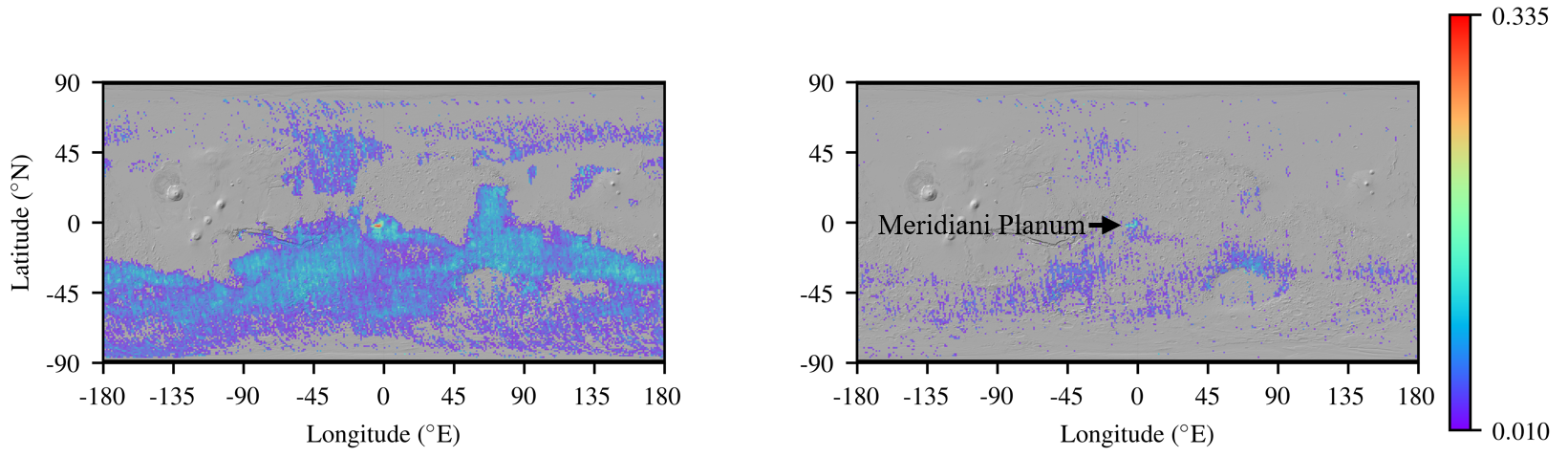}
\caption{\textbf{Martian Surface Map:} The images show the mineral map for hematite of the Martian surface produced by FCLS (left) and analysis-by-synthesis (right) using TES data. Both algorithms find the known deposit of hematite on Meridiani Planum, but analysis-by-synthesis predicts a sparser map which matches expectated distributions.}
\label{fig:TESresults}
\end{figure*}

\textbf{TES Data.} The Mars TES data was unmixed using our analysis-by-synthesis method to demonstrate it's utility on tasks where zero training data is available. The method produces mineral maps which correctly finds abundances of the mineral hematite at Meridiani Planum in Figure~\ref{fig:TESresults}. This is an important Martian mineral which provides evidence for liquid water having existed at some point on Mars, and has been verified by NASA's Opportunity Rover \cite{klingelhofer2004jarosite}. Note how FCLS predicts many sites for hematite, while our method narrows down potential sites on the Martian surface, which is useful for planetary scientists. By allowing for spectral variation through our physics-based approach, our method has lower RMS reconstruction error than previous analysis of TES data. FCLS, which was previously used on TES because of the zero training-data problem, has an average RMS reconstruction error of 0.0043 while analysis-by-synthesis has an average of 0.0038. This is an exciting result as our methods could provide a new suite of hyperspectral analysis tools for scientists studying the Martian surface.

\section{Discussion}

This paper incorporated generative physics models into spectral unmixing algorithms via differentiable programming. We adopt a physics-based dispersion model to simulate spectral variability, and show how this model can realistically fit several real measured spectra via gradient descent. We further show how to jointly optimize for the dispersion parameters and material abundances with an analysis-by-synthesis optimization. A second algorithm is introduced for tasks where additional data is available by training a CNN to ``Inversely Render" a hyperspectral image with the differentiable dispersion model in the loop. 

We validate these contributions extensively on three datasets ranging from mid to far IR and VNIR, and compared against state-of-the-art optimization, statistical and deep learning benchmarks. From these experiments we observe that analysis-by-synthesis has the best performance when training data is not available, and that Inverse Rendering has the best performance when training data is available. We also see that analysis-by-synthesis is noise resilient, and reconstructs Mars spectra with lower error than previous techniques.

There are still limitations for the methods proposed. First, analysis-by-synthesis has a large computational cost compared to other methods, although this could be mitigated through parallelization of the algorithm. Secondly, the spectral unmixing community is limited by the training data available. This is difficult to overcome, because it is expensive to produce quality datasets, and it is not easy for experts to label remote sensing datasets from prior knowledge alone. Future work could investigate generating realistic synthetic data suitable for training machine learning based algorithms for better performance. We hope using generative physics-based models inspires others to produce realistic synthetic data as well as differentiable programming methods which require low training data. 

\textbf{Acknowledgements.} This work was supported by NSF grant IIS-1909192 as well as GPU resources from ASU Research Computing. We thank Dr. Alina Zare, Christopher Haberle, and Dr. Deanna Rogers for their helpful discussions, and Kim Murray (formerly Kim Feely) for providing the laboratory measurements and analysis contributing to this paper.

\clearpage
%
%
\bibliographystyle{splncs04}
\bibliography{egbib.bib}

\newpage

\title{Supplemental Material: Differentiable Programming for Hyperspectral Unmixing using a Physics-based Dispersion Model} 
\author{}
\institute{}

\maketitle

\section*{Appendix A: Derivation of Dispersion Model}
In this section, we derive the dispersion model from first principles, modeling the generation of spectral radiance as a sum of mass-spring oscillations driven by an electromagnetic wave. This induces changes in the index of refraction, which governs the reflectance of the material with respect to light wavelength/frequency. This is based on earlier work by Garbuny and by Spitzer et al.~\cite{garbuny1965optical,spitzer1961infrared}. 

We first start with the equation for a mass-spring oscillator driven by an external force:
\begin{equation}
    F = m\frac{d^2x}{dt^2} + R\frac{dx}{dt} + G\cdot x(t).
\end{equation}

For a charged particle, $F = qE$, where $q$ is charge and $E = E_0 e^{i\omega t}$ for a propagating electromagnetic wave. Thus we can substitute these in to get:
\begin{equation}
    F =q E_0 e^{i\omega t}= m\frac{d^2x}{dt^2} + R\frac{dx}{dt} + G\cdot x(t)
\end{equation}
which has the solution:
\begin{equation}
\label{eq:massspring}
    x = \frac{q E_0 e^{i\omega t}}{m} \frac{1}{\frac{G}{m} - \omega^2 + i \frac{R}{m} \omega} = \frac{q E_0 e^{i\omega t}}{m} \frac{1}{\omega_0^2 - \omega^2 + i \gamma \omega}.
\end{equation}

where $\omega_0^2 = G/m$ and $\gamma = R/m.$ 

At the same time, we can also relate $x$ to $E$ via the band strength:
\begin{equation}
    x = \frac{\alpha E}{q}
\end{equation}
where $\alpha$ is the polarizability. Using the identity $\epsilon = 1 + 4\pi \alpha$, we can derive the following band strength equation:
\begin{equation}
\label{eq:bandstrength}
    x = \frac{(\epsilon -1) E_0 e^{i\omega t}}{4\pi q}.
\end{equation}

Combining Eq.~\ref{eq:massspring} and Eq.~\ref{eq:bandstrength}, we get 
\begin{equation}
    \frac{(\epsilon -1) E_0 e^{i\omega t}}{4\pi q} = \frac{q E_0 e^{i\omega t}}{m} \frac{1}{\omega_0^2 - \omega^2 + i \gamma \omega}.
\end{equation}
Solving for $\epsilon$:
\begin{equation}
   \epsilon = \frac{4 \pi q^2}{m} \frac{1}{\omega_0^2 - \omega^2 + i \gamma \omega} + 1.
\end{equation}

Relating $\epsilon$ to the refractive index, 
$\epsilon \mu = \hat{n}^2$ where $\mu = 1$ and $\hat{n} = n - ik$, we get 
\begin{equation}
    \epsilon = (n-ik)^2 = n^2 - k^2 - 2nki = \frac{4 \pi q^2}{m} \frac{1}{\omega_0^2 - \omega^2 + i \gamma \omega} + 1
\end{equation}
This yields the refractive index equations:
\begin{equation}
    n^2 - k^2 = \frac{4 \pi q^2}{m} \frac{\omega_0^2 - \omega^2}{(\omega_0^2 - \omega^2)^2 + \gamma^2 \omega^2} + 1
\end{equation}
\begin{equation}
    2nk = \frac{4 \pi q^2}{m} \frac{\gamma \omega}{(\omega_0^2 - \omega^2)^2 + \gamma^2 \omega^2}.
\end{equation}


Using the Lorentz-Lorenz formula, we can get
\begin{equation}
    \hat{n}^2 = 1 + \frac{4 \pi q^2}{m} \frac{1}{\omega_1^2 - \omega^2 + i \gamma \omega}
\end{equation}
where $\omega_1^2 = \omega_0^2 - \frac{4\pi q^2}{3m}$ where $\omega_1 < \omega_0.$ So plugging in $\omega_1$ for $\omega_0$ yields:

\begin{equation}
    n^2 - k^2 = \frac{4 \pi q^2}{m} \frac{\omega_1^2 - \omega^2}{(\omega_1^2 - \omega^2)^2 + \gamma^2 \omega^2} + 1
\end{equation}
\begin{equation}
    2nk = \frac{4 \pi q^2}{m} \frac{\gamma \omega}{(\omega_1^2 - \omega^2)^2 + \gamma^2 \omega^2}.
\end{equation}

This entire derivation was for a single oscillator, but in practice, there are multiple oscillators that interact. We write this as a linear superposition given as follows:

\begin{equation}
    n^2 - k^2 = \epsilon_r + \sum_i \frac{4 \pi q^2 f_i}{m_i} \frac{\omega_i^2 - \omega^2}{(\omega_i^2 - \omega^2)^2 + \gamma_i^2 \omega^2}
\end{equation}
\begin{equation}
    2nk = \sum_i \frac{4 \pi q^2 f_i}{m_i} \frac{\gamma_i \omega}{(\omega_i^2 - \omega^2)^2 + \gamma_i^2 \omega^2}.
\end{equation}

where $f_i$ is the strength of each individual oscillator. Using these equations, we have two equations for two unknowns ($n$ and $k$), which we showed in Section 3 of the main paper is the basis of calculating reflectance and emission. 

\section*{Appendix B: Spectral Variation}

As stated in the main paper, the motivation for incorporating the dispersion model into a differentiable program for spectral unmixing is to allow for physically plausible spectral variation of pure materials. Because it is known that absorption bands can shift in frequency and strength between different endmember samples, the goal was to find a model that described these changes in a physically plausible way. That is we wanted a generative model for the formation of spectra with parameters that have "dials" to tune the frequency, strength, and shape of absorption bands. From the literature on analysis of the formation of spectra from an atomistic perspective \cite{spitzer1961infrared,wenrich1996optical} we find that the Lorentz-Lorenz dispersion model is the correct approach to take. However, unlike previous works we go further than using the model to derive optical properties of materials, we also incorporate the dispersion model into an end-to-end spectral unmixing pipeline that allows the parameters to be fine-tuned via differentiable programming to account for spectral variability.

In Appendix A, the dispersion model is derived from first principles and each absorption band is described by the parameters $\bm{\rho},\bm{\omega_o}, \bm{\gamma}, \bm{\epsilon_r}$. $\bm{\rho}$ is the band strength and as it increases the absorption band becomes deeper. $\bm{\omega_o}$ is the resonant frequency and as it increases the absorption band shifts in wavenumber (and also slightly shifts the shape). $\bm{\gamma}$ is the frictional force (dampening coefficient) and controls the shape/width of the absorption bands. $\bm{\epsilon_r}$ is relative dielectric permeability and as it increases the entire emissivity curve is shifted downwards. Also note that absorption bands which are close to each other interact in highly non-linear ways.

\begin{figure}[h!]
    \centering
    \includegraphics[width=0.49\textwidth]{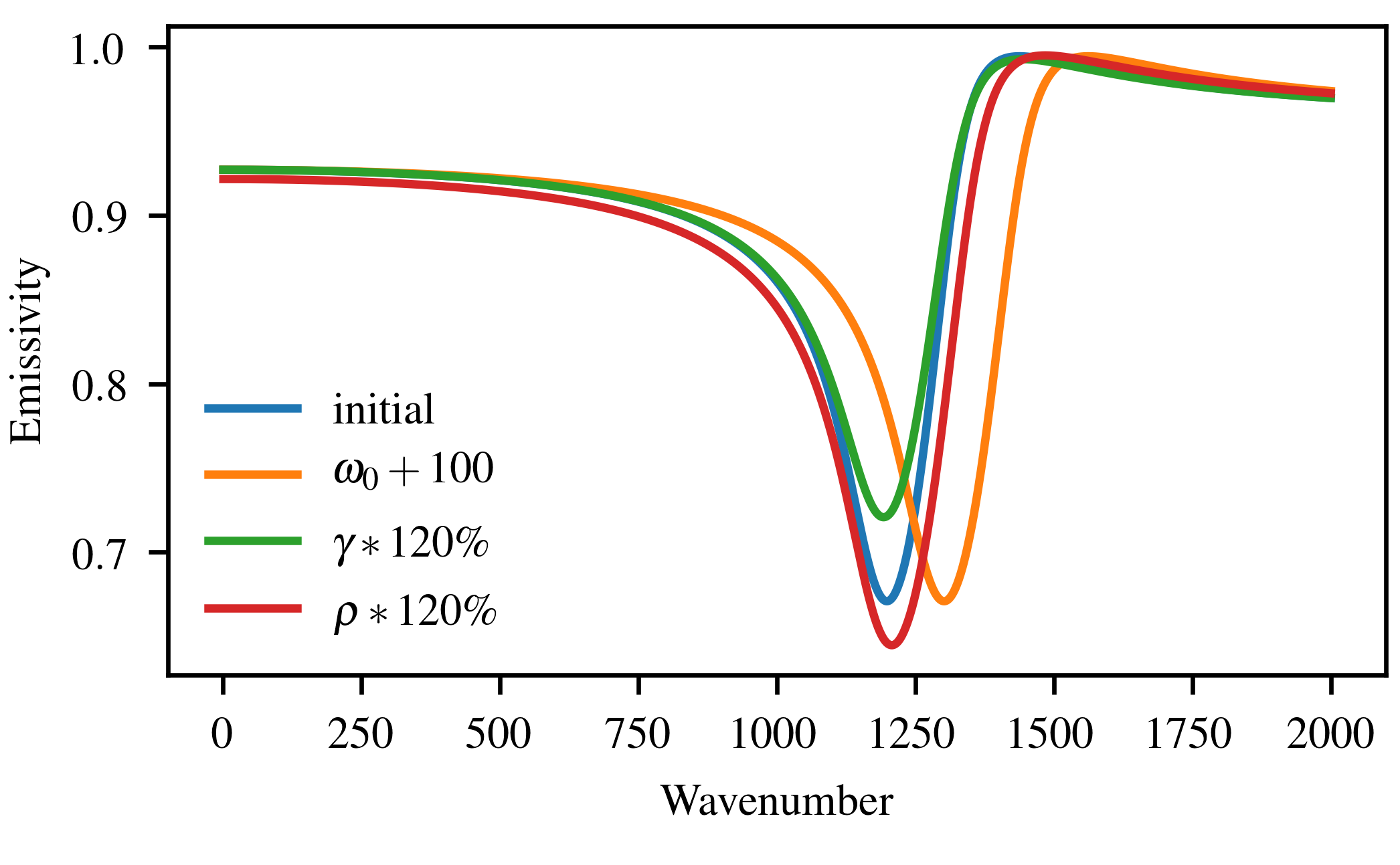}
    \caption{A single absorption band is initialized with $\epsilon_r=2.356, \omega_0=1161, \gamma=0.1, \rho=0.67$. Then the parameters are perturbed such that $\omega_0$ is increased by 100, $\gamma$ is increased to 120\%, and $\rho$ is increased to 120\%. The plots show the effect of changing each parameter individually to show it's control over the shape and width of the absorption band.}
    \label{fig:my_label}
\end{figure}

The importance of initializing the alternating optimization with good initial dispersion parameters was emphasized in the main paper, as the problem is non-convex and good initialization is essential. It also makes intuitive sense to initialize with parameters fit to an endmember sample to give physical significance to the generated spectra. As shown in the results of the main paper, we achieve good fits with low MSE on endmember libraries used to analyze the Martian surface as well as a semi-urban university scene. The endmember libraries used to fit the minerals to analyze the Mars TES data are of high quality from the Arizona State University Thermal Emission Spectral Library \cite{christensen2000thermal}. The resulting parameters from a few of the important materials from this endmember library are provided in the following tables.

\begin{table}[htbp]
  \centering
  \caption{Dispersion parameters found for Olivine Fo10}
  \small
    \begin{tabular}{P{1.3cm}P{1.3cm}P{1.3cm}P{1.3cm}P{1.3cm}P{1.3cm}}
    \hline
    \multicolumn{1}{c}{Axis} & \multicolumn{1}{c}{Index} & \multicolumn{1}{c}{$\omega_0$} & \multicolumn{1}{c}{$\gamma$} & \multicolumn{1}{c}{$\rho$} & \multicolumn{1}{c}{$\epsilon_r$} \\
    \hline

    0     & 0     & 258.45 & 0.018 & 0.022 & 1.07 \\
    0     & 1     & 272.71 & 0.038 & 0.070 & 1.07 \\
    0     & 2     & 285.33 & 0.027 & 0.035 & 1.07 \\
    0     & 3     & 340.81 & 0.021 & 0.015 & 1.07 \\
    0     & 4     & 361.06 & 0.067 & 0.187 & 1.07 \\
    0     & 5     & 467.03 & 0.060 & 0.091 & 1.07 \\
    0     & 6     & 589.36 & 0.032 & 0.043 & 1.07 \\
    0     & 7     & 826.60 & 0.011 & 0.015 & 1.07 \\
    0     & 8     & 863.05 & 0.030 & 0.083 & 1.07 \\
    0     & 9     & 934.94 & 0.018 & 0.038 & 1.07 \\
    0     & 10    & 1068.56 & 0.009 & 0.001 & 1.07 \\
    0     & 11    & 1349.50 & 0.043 & 0.009 & 1.07 \\
    0     & 12    & 1400.46 & 0.057 & 0.026 & 1.07 \\
    0     & 13    & 1452.82 & 0.064 & 0.020 & 1.07 \\
    0     & 14    & 1518.96 & 0.079 & 0.025 & 1.07 \\
    0     & 15    & 1597.62 & 0.018 & 0.001 & 1.07 \\
    0     & 16    & 1694.56 & 0.043 & 0.007 & 1.07 \\
    0     & 17    & 1794.69 & 0.032 & 0.002 & 1.07 \\
    0     & 18    & 1837.96 & 0.009 & 0.001 & 1.07 \\
    0     & 19    & 1934.50 & 0.056 & 0.020 & 1.07 \\
    1     & 0     & 293.77 & 0.042 & 0.240 & 1.99 \\
    1     & 1     & 303.28 & 0.058 & 0.263 & 1.99 \\
    1     & 2     & 317.16 & 0.137 & 0.356 & 1.99 \\
    1     & 3     & 473.47 & 0.006 & 0.002 & 1.99 \\
    1     & 4     & 496.39 & 0.029 & 0.030 & 1.99 \\
    1     & 5     & 504.45 & 0.062 & 0.302 & 1.99 \\
    1     & 6     & 562.92 & 0.055 & 0.057 & 1.99 \\
    1     & 7     & 577.32 & 0.027 & 0.008 & 1.99 \\
    1     & 8     & 891.85 & 0.023 & 0.189 & 1.99 \\
    1     & 9     & 990.28 & 0.047 & 0.086 & 1.99 \\
    1     & 10    & 1108.25 & 0.023 & 0.006 & 1.99 \\
    \end{tabular}%
  \label{tab:olivineFo10}%
\end{table}%

\begin{table}[htbp]
  \centering
  \caption{Dispersion parameters found for Biotite}
   \small
    \begin{tabular}{P{1.3cm}P{1.3cm}P{1.3cm}P{1.3cm}P{1.3cm}P{1.3cm}}
    \hline
    \multicolumn{1}{c}{Axis} & \multicolumn{1}{c}{Index} & \multicolumn{1}{c}{$\omega_0$} & \multicolumn{1}{c}{$\gamma$} & \multicolumn{1}{c}{$\rho$} & \multicolumn{1}{c}{$\epsilon_r$} \\
    \hline
    0     & 0     & 235.91 & 0.066 & 0.2343 & 1.31 \\
    0     & 1     & 432.39 & 0.056 & 0.4040 & 1.31 \\
    0     & 2     & 439.80 & 0.039 & 0.4131 & 1.31 \\
    0     & 3     & 446.34 & 0.014 & 0.0385 & 1.31 \\
    0     & 4     & 451.92 & 0.042 & 0.4797 & 1.31 \\
    0     & 5     & 594.57 & 0.073 & 0.0147 & 1.31 \\
    0     & 6     & 954.50 & 0.036 & 0.2510 & 1.31 \\
    0     & 7     & 1008.94 & 0.014 & 0.0578 & 1.31 \\
    0     & 8     & 1013.39 & 0.017 & 0.0184 & 1.31 \\
    0     & 9     & 1041.20 & 0.048 & 0.0178 & 1.31 \\
    0     & 10    & 1075.68 & 0.025 & 0.0198 & 1.31 \\
    0     & 11    & 1116.66 & 0.007 & 0.0003 & 1.31 \\
    0     & 12    & 1152.61 & 0.019 & 0.0012 & 1.31 \\
    0     & 13    & 1390.98 & 0.044 & 0.0177 & 1.31 \\
    0     & 14    & 1460.91 & 0.061 & 0.0280 & 1.31 \\
    0     & 15    & 1524.44 & 0.065 & 0.0676 & 1.31 \\
    0     & 16    & 1629.72 & 0.025 & 0.0271 & 1.31 \\
    0     & 17    & 1661.44 & 0.007 & 0.0034 & 1.31 \\
    0     & 18    & 1687.84 & 0.068 & 0.0723 & 1.31 \\
    0     & 19    & 1772.30 & 0.074 & 0.0877 & 1.31 \\
    0     & 20    & 1813.27 & 0.006 & 0.0009 & 1.31 \\
    0     & 21    & 1865.48 & 0.064 & 0.0731 & 1.31 \\
    0     & 22    & 1964.44 & 0.055 & 0.0131 & 1.31 \\
    1     & 0     & 268.77 & 0.073 & 0.4634 & 2.61 \\
    1     & 1     & 294.51 & 0.045 & 0.1965 & 2.61 \\
    1     & 2     & 313.92 & 0.064 & 0.3242 & 2.61 \\
    1     & 3     & 337.12 & 0.093 & 0.4930 & 2.61 \\
    1     & 4     & 362.24 & 0.062 & 0.1954 & 2.61 \\
    1     & 5     & 400.00 & 0.209 & 0.5174 & 2.61 \\
    1     & 6     & 462.66 & 0.065 & 0.4399 & 2.61 \\
    1     & 7     & 492.95 & 0.080 & 0.3498 & 2.61 \\
    1     & 8     & 510.47 & 0.061 & 0.0664 & 2.61 \\
    1     & 9     & 653.21 & 0.078 & 0.0611 & 2.61 \\
    1     & 10    & 718.49 & 0.040 & 0.0331 & 2.61 \\
    1     & 11    & 873.68 & 0.115 & 0.3343 & 2.61 \\
    1     & 12    & 928.32 & 0.048 & 0.0488 & 2.61 \\
    1     & 13    & 991.97 & 0.015 & 0.3550 & 2.61 \\
    1     & 14    & 1588.86 & 0.040 & 0.0607 & 2.61 \\
    1     & 15    & 1963.15 & 0.004 & 0.0023 & 2.61 \\
    1     & 16    & 1989.53 & 0.001 & 0.0002 & 2.61 \\
    \end{tabular}%
  \label{tab:biotiteParams}%
\end{table}%

\begin{table}[htbp]
  \centering
  \caption{Dispersion parameters found for Hematite}
    \small
    \begin{tabular}{P{1.3cm}P{1.3cm}P{1.3cm}P{1.3cm}P{1.3cm}P{1.3cm}}
    \hline
    \multicolumn{1}{c}{Axis} & \multicolumn{1}{c}{Index} & \multicolumn{1}{c}{$\omega_0$} & \multicolumn{1}{c}{$\gamma$} & \multicolumn{1}{c}{$\rho$} & \multicolumn{1}{c}{$\epsilon_r$} \\
    \hline
    0     & 0     & 258.29 & 0.11  & 0.110 & 1.27 \\
    0     & 1     & 279.35 & 0.13  & 0.141 & 1.27 \\
    0     & 2     & 294.73 & 0.11  & 0.149 & 1.27 \\
    0     & 3     & 335.86 & 0.08  & 0.130 & 1.27 \\
    0     & 4     & 471.32 & 0.07  & 0.098 & 1.27 \\
    0     & 5     & 526.58 & 0.05  & 0.029 & 1.27 \\
    0     & 6     & 543.94 & 0.07  & 0.062 & 1.27 \\
    0     & 7     & 563.14 & 0.08  & 0.067 & 1.27 \\
    0     & 8     & 609.37 & 0.04  & 0.041 & 1.27 \\
    0     & 9     & 619.61 & 0.04  & 0.041 & 1.27 \\
    0     & 10    & 632.43 & 0.07  & 0.067 & 1.27 \\
    0     & 11    & 654.46 & 0.09  & 0.054 & 1.27 \\
    0     & 12    & 686.74 & 0.12  & 0.038 & 1.27 \\
    0     & 13    & 798.98 & 0.04  & 0.011 & 1.27 \\
    0     & 14    & 890.21 & 0.03  & 0.009 & 1.27 \\
    0     & 15    & 916.82 & 0.02  & 0.005 & 1.27 \\
    0     & 16    & 958.26 & 0.04  & 0.014 & 1.27 \\
    0     & 17    & 1002.55 & 0.04  & 0.010 & 1.27 \\
    0     & 18    & 1100.72 & 0.03  & 0.022 & 1.27 \\
    0     & 19    & 1167.07 & 0.02  & 0.010 & 1.27 \\
    0     & 20    & 1238.37 & 0.01  & 0.005 & 1.27 \\
    0     & 21    & 1282.36 & 0.03  & 0.019 & 1.27 \\
    1     & 0     & 234.31 & 0.02  & 0.007 & 1.25 \\
    1     & 1     & 238.56 & 0.06  & 0.031 & 1.25 \\
    1     & 2     & 312.13 & 0.09  & 0.255 & 1.25 \\
    1     & 3     & 356.47 & 0.04  & 0.032 & 1.25 \\
    1     & 4     & 430.53 & 0.09  & 0.085 & 1.25 \\
    1     & 5     & 444.75 & 0.06  & 0.032 & 1.25 \\
    1     & 6     & 457.95 & 0.04  & 0.011 & 1.25 \\
    1     & 7     & 486.07 & 0.03  & 0.019 & 1.25 \\
    1     & 8     & 577.56 & 0.08  & 0.160 & 1.25 \\
    1     & 9     & 727.69 & 0.06  & 0.049 & 1.25 \\
    1     & 10    & 748.13 & 0.07  & 0.040 & 1.25 \\
    1     & 11    & 773.90 & 0.06  & 0.013 & 1.25 \\
    1     & 12    & 1049.92 & 0.10  & 0.058 & 1.25 \\
    1     & 13    & 1069.60 & 0.01  & 0.003 & 1.25 \\
    1     & 14    & 1140.36 & 0.02  & 0.012 & 1.25 \\
    1     & 15    & 1197.28 & 0.04  & 0.022 & 1.25 \\
    1     & 16    & 1256.54 & 0.02  & 0.010 & 1.25 \\
    \end{tabular}%
  \label{tab:hematiteParams}%
\end{table}%

\section*{Appendix C: Alternating Optimization Convergence}\label{app:alt-min}

Our ultimate goal is to solve the spectral unmixing problem which can be formulated as $\min_{\mathbf{A}, \mathbf{x}}\left\| \mathbf{b} - \mathbf{A} \mathbf{x}\right\|_2^2$, where the minimization occurs over both  the matrix $\mathbf{A}$ and the unmixing vector $\mathbf{x}$. This is a standard case of alternate minimization which is known to be nonconvex~\cite{jain2017non}. In practice, alternating minimzation are particularly hard to tackle due to the presence of suboptimal local minimas. Recent progress on tackling nonconvex problems involves either characterizing the optimization landscape~\cite{ge2017no,thaker2020sample,Sun_2017} or providing initialization to descent algorithms~\cite{bhojanapalli2016dropping,chen2019gradient} to assure convergence to the global optimum. It is known that gradient descent applied to alternate minimization problem faces the issue of getting stuck at local minimas~\cite{jain2017non} and hence initialization plays an important role in solving Equation 7 in the main paper. With that in mind, in this paper we provide a mechanism to provide good initialization to gradient descent algorithm with the hope of tackling the alternating minimization problem effectively. 
\\
\\
\noindent
\textbf{Initialization using dictionary $A(\epsilon^{\text{model}})$}: We investigate the properties of matrix $A$ as relates to the convergence of the alternating optimization. We denote the measured the emissivity spectrum of various materials in the lab as $\epsilon^{\text{measured}}$, and the physics-based dispersion model as $\epsilon^{\text{model}}$. We then use these emissivity spectrum to construct a dictionary $A(\epsilon)$ which servers as the initialization for $A$ in the alternating minimization approach in Equation 7 of the main paper. The intuition behind this revolves around the ability of matrix A as a dictionary of known emissivity spectra and we expect that the unknown spectra $\epsilon^{\text{unknown}}$ would be described as a linear combination of columns from matrix $A$. 

Consider the following subproblem of the alternating minimization:
\begin{align}
    \min_{\mathbf{x}}\|\mathbf{b} - \mathbf{A}(\Lambda)\mathbf{x}\|_2^2\nonumber
\end{align} without the regularization terms. In order to ensure the uniqueness of the solution $\mathbf{x}^*$, we need to ensure that the matrix $A$ is full rank. The rate of convergence for the above minimization is inversely dependent on the condition number of the matrix $\mathbf{A}(\Lambda)$. While it is difficult to analyze this matrix analytically, we perform an experimental characterization of the rank, condition number, and eigenvalues of the matrix for several different runs of the optimization algorithm with random initializations.  


\begin{figure}[h!]
    \centering
    \begin{subfigure}{0.49\textwidth}
        \centering
        \includegraphics[height=0.6\textwidth]{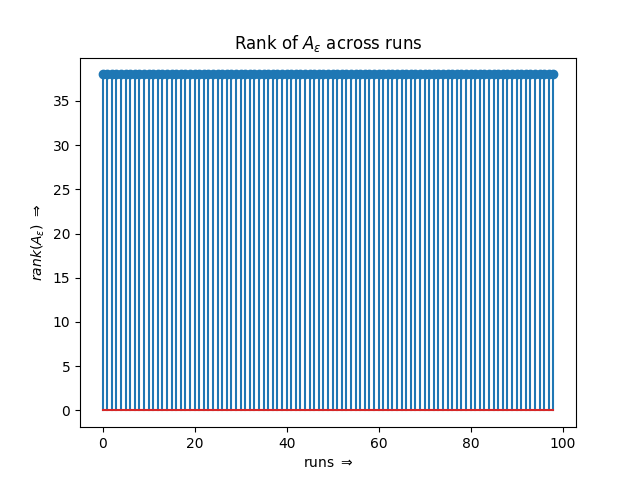}
        \caption{Rank of $\mathbf{A}$ across runs}
    \end{subfigure}%
    \hfill
    \begin{subfigure}{0.49\textwidth}
        \centering
        \includegraphics[height=0.6\textwidth]{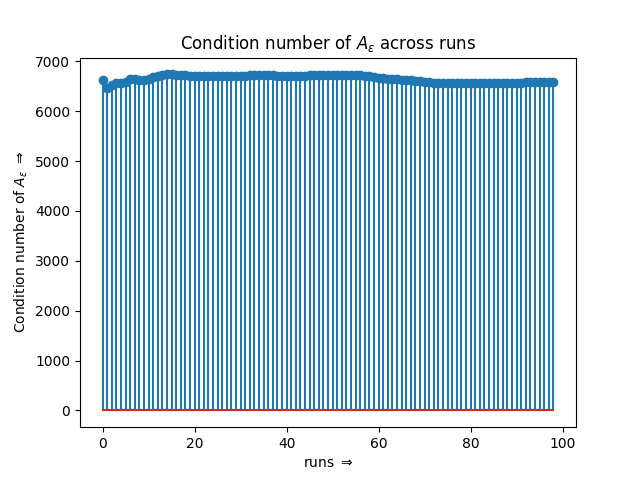}
        \caption{Condition number of $\mathbf{A}$ across runs}
    \end{subfigure}
    \vskip\baselineskip
    \centering
    \begin{subfigure}{0.49\textwidth}
        \centering
        \includegraphics[height=0.6\textwidth]{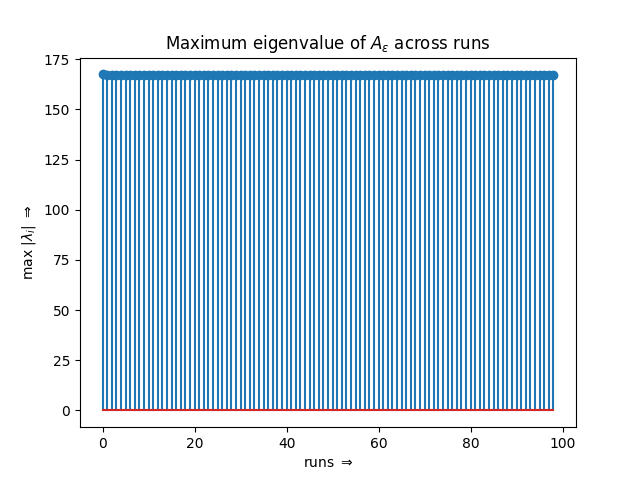}
        \caption{Max eigenvalue of $\mathbf{A}$ across runs}
    \end{subfigure}%
    \hfill
    \begin{subfigure}{0.49\textwidth}
        \centering
        \includegraphics[height=0.6\textwidth]{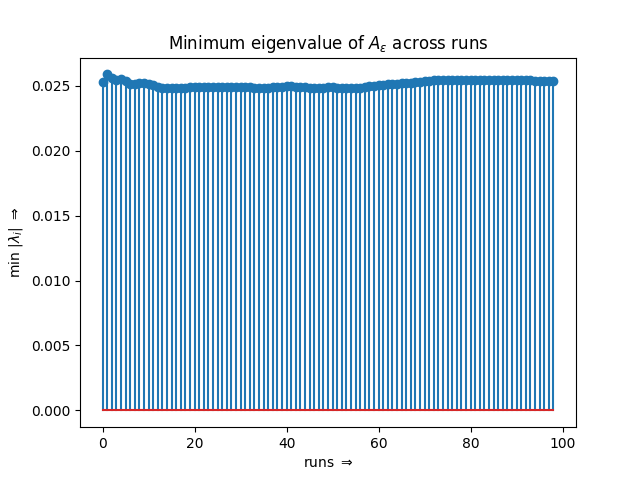}
        \caption{Min eigenvalue of $\mathbf{A}$ across runs}
    \end{subfigure}
    \caption{Behaviour of $\mathbf{A}$ across runs}
\end{figure}

\begin{figure}[h!]
    \centering
    \includegraphics[height=0.49\textwidth]{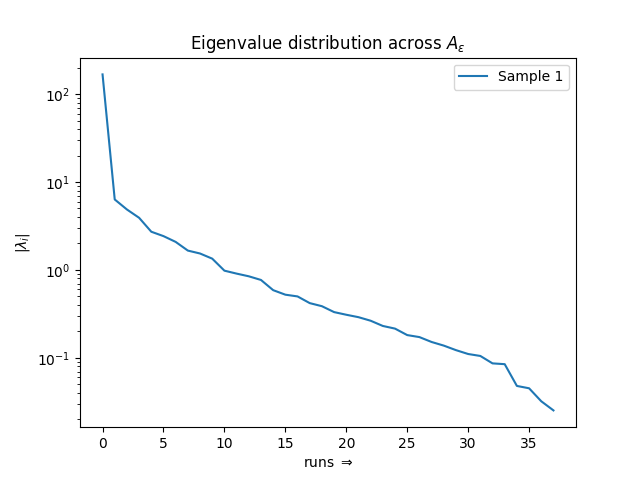}
    \caption{Random sample of $\mathbf{A}$ has the following eigenvalue distribution}
    \label{fig:my_label}
\end{figure}

From the plots, we can note that the matrix $\mathbf{A}(\Lambda)$ has full rank with condition number of around 6000. The minimum and maximum eigenvalues are not showing drastic difference which goes well with our motive to incorporate small changes using alternate minimization to fit the spectral differences due to geographic differences. The high condition number is the reason for the relatively slow performance for running the alternating minimization framework, with our method taking tens of seconds to converge. 

\section*{Appendix D: Implementation Details}

\textbf{Fully Constrained Least Squares} We compare against Fully Constrained Least Squares (FCLS) ~\cite{heinz2001fully} which is a popular classical unmixing algorithm. FCLS solves for the the (aerial) abundances, $\mathbf{x}$: $ \hat{\mathbf{x}} = \argmin_{\mathbf{x}} ||\mathbf{b} - \mathbf{A}\mathbf{x}||_2^2$ subject to $  ||\mathbf{x}||_1 = 1,  \mathbf{x} \geq 0$. The constraints, referred to as the sum-to-one constraint and the non-negativity constraint, are enforced since abundances are interpreted as percentages.
   
\textbf{Normal Compositional Model.} The Normal Compositional Model (NCM) \cite{stein2003application} is one of the most popular methods for modelling endmember variability via statistical methods. The method requires a small amount of training data (roughly 50 samples per endmember) to learn the mean and variance of reflectivity (or emissivity) of each spectral band, and modelling the variation as a Gaussian distribution. During unmixing, Expectation Maximization is used to simultaneously learn the abundances and the endmember variation, subject to the abundance sum-to-one and non-negativity constraints. 

The NCM is run using the Matlab code provided by Du et al.~\cite{du2014spatial}. Training data of about 50 samples of endmember variation were provided to the NCM for each dataset. There are no hyperparameters needed for this method.

\textbf{Beta Compositional Model.} The Beta Compositional Model (BCM) \cite{du2014spatial} is a more recent method for modelling spectral variability via a statistical method. Similar to the NCM, a small amount of training data is used to learn the beta parameters of each spectral band, and an Expectation Maximization algorithm is used during unmixing. The beta parameters allow each spectral band to be modelled as a more complex distribution than the NCM and has been shown to increase performance. 

The BCM is run using the Matlab code provided by Du et al.~\cite{du2014spatial}. Training data of about 50 samples of endmember variation were provided to the BCM for each dataset. For datasets without sufficient endmember samples, we generated synthetic endmember variation with the dispersion model. We search for the optimal hyperparameters through repeated experiments and report the best results. The optimal BCM across all datasets was run with K = 3, $\sigma_V$ = 100, and $\sigma_M$ = 0.001.

\textbf{CNN for Spectral Classification and Unmixing.} We baseline against the architecture recently proposed by Zhang et al.~\cite{zhang2018spectral} for hyperspectral unmixing using both a 1D and 3D CNN. The main difference between the 1D and 3D CNN is that the 3D CNN is operates on $3\times 3 $ bundles of pixels while the 1D CNN is provided a single pixel. However, in both architectures the convolutional kernal is 1D and operates along the spectral dimension. Both architectures have four convolutional layers with Rectified Linear Unit (ReLU) and max-pooling non-linear operations. These layers are followed by 2 fully connected layers, where the last layer is the output abundance predictions. The ReLU of the fully connected layers ensure non-negativity, and the sum-to-one constraint is ensured by normalizing the output layer. The network is trained to minimize $-\hat{\mathbf{x}} \log(\mathbf{x})$, where $\hat{\mathbf{x}}$ and  $\mathbf{x}$ are the predicted and ground truth abundances respectively.  While the 1D architecture performs well on the dataset it was designed for in~\cite{zhang2018spectral}, we found that modifications were necessary to maximize the performance on our datasets. Namely, we found that training the network with respect to the mean squared error (MSE) loss function, removing the max-pooling layers, and adding an additional fully connected layer before the output improved performance of the 1D CNN. The 3D CNN is only applicable to datasets with spatial information, thus is ignored for the Feely and Synthetic datasets.

\textbf{CNN Baselines~\cite{zhang2018spectral}.} The architecture from Zhang et al. was used to baseline against our method (CNN-1D and CNN-3D), as well as introducing our own modified baseline (CNN-1D modified). All CNNs were trained for 100 epochs using the Adam optimizer with learning = 0.0005, betas = (0.9, 0.999) and weight decay = 0.  
\\
\textit{CNN-1D:} A 1x1 hyperspectral pixel is input into the network, four convolutional layers with alternating 1x5 and 1x4 kernels and a depth of 3, 6, 12, and 24 kernels in each layer respectively. All convolutional layers have ReLU activations a 1x2 maxpooling layer.  The convolutional layers are followed by a fully connected layer with 150 hidden units, and an output fully connected layer with a size that depends on the number of abundances. Normalization is used to enforce the abundance sum-to-one constraint and the ReLU activation enforces the non-negativity constraint. The CNN is trained to minimize the log loss between the predicted and ground truth abundances. The network converges in about 100 epochs with a learning rate of 1e-3.
\\
\textit{CNN-3D:} CNN-3D has an almost identical architecture, although it accepts a 3x3 set of pixels at the input. Although a spatial dimension exists at the input, the convolutions only occur in the spectral dimension. four convolutional layers with alternating 1x5 and 1x4 kernels and a depth of 16, 32, 64, and 128 kernels in each layer respectively.  The convolutional layers are followed by a fully connected layer with 150 hidden units, and an output fully connected layer with a size that depends on the number of abundances. Normalization is used to enforce the abundance sum-to-one constraint and the ReLU activation enforces the non-negativity constraint. The CNN is trained to minimize the log loss between the predicted and ground truth abundances. The network converges in about 100 epochs with a learning rate of 1e-3.
\\
\textit{CNN-1D Modified:} Finally, a modified version of CNN-1D is baselined against to try to find the optimal architecture for performance on our datasets. The first 2 max-pooling layers are removed, an additional hidden fully connected layer with 150 units is added before the output, and a softmax operation is applied to the output to enforce the abundance sum-to-one constraint. Also, the network is trained to minimize the mean squared error between the predicted and ground truth abundances. The network converges in about 100 epochs with a learning rate of 1e-3.

\textbf{Analysis-by-Synthesis Optimization.} For analysis-by-synthesis, the sparse regularization was set with p = 0.95 and $\lambda_p$ = 0.0001. Dispersion parameters were constrained within a tolerance of their initial conditions with $\rho_{tol}$  = 0.05, $\gamma_{tol}$ = 0.005, $\epsilon_{tol}$ = 0.001, and $\omega_{tol}$ = 0.0001. On the Gulfport datasets $\gamma_{tol}$ and $\epsilon_{tol}$ were increased to 0.05 to compensate for increased variation. Analysis-by-Synthesis alternates between finding optimal abundances (solving a regularized least squares problem), and updating the dispersion parameters for 100 iterations using the Adam optimizer with learning rate = 0.01, betas = (0.9, 0.999), and weight decay = 0.

\textbf{Inverse Rendering CNN.} The inverse rendering CNN uses the same CNN architecture as CNN-1D modified. The input to the CNN is the spectrum (or batch of spectra). That is there are four convolutional layers with alternating 1x5 and 1x4 kernels and a depth of 3, 6, 12, and 24 kernels in each layer respectively. The convolutional layers also have ReLU activations and the last 2 layers have a 1x2 maxpooling layer.  The convolutional layers are followed by fully connected layers with 150 hidden units. The final fully connected layer has enough units for the amount of dispersion parameters and abundances depending on the size of the endmember library and number of dispersion parameters per endmember. Then, the dispersion parameters are used to render endmember spectra and the mixture is reconstructed under the linear mixing model with the predicted abundances as inputs. The network only needs the input spectra and the abundances as inputs for training, as the reconstruction error of the spectra is used to back-propagated through the differentiable dispersion model to teach the network to predict good dispersion parameters. Real data (when available) and synthetic data (around 50,000 samples) are used to train the network, which converges after about 100 epochs. An Adam optimizer is used with learning rate set to 1e-3, betas set to (0.9, 0.999), and weight decay set to 0.

\section*{Appendix E: Additional Results} Due to size limitations on ArXiv, we cannot display all the mineral maps of Mars TES data here. We encourage the reader to go to Dr. Jayasuriya's webpage \url{https://web.asu.edu/sites/default/files/imaging-lyceum/files/eccv2020_hyperspectral_paperwithsupplement-min.pdf} to see the full supplemental material.

%
%
%
%
%
%
%
%
%
%
%

\clearpage
%
%
\bibliographystyle{splncs04}
\bibliography{egbib.bib}
\end{document}